\documentclass[lettersize,journal]{IEEEtran}
\usepackage{amsmath,amsfonts}
\usepackage{algorithmic}
\usepackage{algorithm}
\usepackage{array}
\usepackage[caption=false,font=normalsize,labelfont=sf,textfont=sf]{subfig}
\usepackage{textcomp}
\usepackage{tabularx}
\usepackage{stfloats}
\usepackage{url}
\usepackage{verbatim}
\usepackage{graphicx}
\usepackage{cite}
\usepackage{lscape}
\usepackage{authblk}
\usepackage{hyperref}
\usepackage{xcolor}
\begin{document}

\title{CerraData-4MM: A multimodal benchmark dataset on Cerrado for land use and land cover classification}
\author[a]{Mateus de Souza Miranda}
\author[b]{Ronny H\"ansch}
\author[a]{Valdivino Alexandre de Santiago J{\'u}nior\thanks{CONTACT Valdivino Alexandre de Santiago J{\'u}nior. Email: valdivino.santiago@inpe.br}}
\author[a]{Thales Sehn K{\"o}rting}
\author[a,c]{Erison Carlos dos Santos Monteiro}
\affil[a]{Laborat\'orio de Intelig\^{e}ncia ARtificial para Aplica\c{c}\~oes AeroEspaciais e Ambientais (LIAREA), Coordena\c{c}\~ao de Pesquisa Aplicada e Desenvolvimento Tecnol\'ogico (COPDT), Instituto Nacional de Pesquisas Espaciais (INPE), S\~ao Jos\'e dos Campos, Brazil}
\affil[b]{German Aerospace Center (DLR), Oberpfaffenhofen, Germany}
\affil[c]{German Society for International Cooperation (GIZ), Bras\'ilia, Brazil}
\markboth{Journal of \LaTeX\ Class Files,~Vol.~14, No.~8, August~2021}%
{Shell \MakeLowercase{\textit{et al.}}: A Sample Article Using IEEEtran.cls for IEEE Journals}
\maketitle

\begin{abstract}
The \textit{Cerrado} faces increasing environmental pressures, necessitating accurate land use and land cover (LULC) mapping despite challenges such as class imbalance and visually similar categories. To address this, we present CerraData-4MM, a multimodal dataset combining Sentinel-1 Synthetic Aperture Radar (SAR) and Sentinel-2 MultiSpectral Imagery (MSI) with 10~m spatial resolution. The dataset includes two hierarchical classification levels with 7 and 14 classes, respectively, focusing on the diverse \textit{Bico do Papagaio} ecoregion. 
We highlight CerraData-4MM's capacity to benchmark advanced semantic segmentation techniques by evaluating a standard U-Net and a more sophisticated Vision Transformer (ViT) model. The ViT achieves superior performance in multimodal scenarios, with the highest macro F1-score of 57.60\% and a mean Intersection over Union (mIoU) of 49.05\% at the first hierarchical level. Both models struggle with minority classes, particularly at the second hierarchical level, where U-Net{’}s performance drops to an F1-score of 18.16\%. Class balancing improves representation for underrepresented classes but reduces overall accuracy, underscoring the trade-off in weighted training. CerraData-4MM offers a challenging benchmark for advancing deep learning models to handle class imbalance and multimodal data fusion. Code, trained models, and data are publicly available at \url{https://github.com/ai4luc/CerraData-4MM}.
\end{abstract}


\section{Introduction}

\begin{table*}[!ht]
\label{tab:table1}
\caption{Related datasets.}
\raggedright
\begin{tabular}{>{\raggedright\arraybackslash}m{2.3cm}>{\raggedright\arraybackslash}m{1.3cm}>{\raggedright\arraybackslash}m{1.3cm}>{\raggedright\arraybackslash}m{1.8cm}>{\raggedright\arraybackslash}m{1cm}>{\raggedright\arraybackslash}m{1cm}>{\raggedright\arraybackslash}m{1cm}>{\raggedright\arraybackslash}m{1cm}>{\raggedright\arraybackslash}m{1.3cm}>{\raggedright\arraybackslash}m{2cm}}
\hline
\textbf{Dataset}&\textbf{Nr. of images}&\textbf{Image size} & \textbf{Spatial resolution (m)} & \textbf{Time series} & \textbf{SAR} & \textbf{Optical} & \textbf{Others} & \textbf{Nr. of classes} & \textbf{Region} \\
\hline
SEN12MS~\cite{SEN12MS}&541,986 & 256$\times$256 & 10 & No & Yes & MSI & -  & 33 & Globally distributed\\
Brazilian \textit{Cerrado}-Savanna Scenes~\cite{nogueira} & 1,311 & 64$\times$64 & 5 & No & No & NIRGB & - & 4 & \textit{Serra do Cipó region}, on \textit{Cerrado} biome  \\
MultiSenGE \cite{MultiSenGE} & 8,157 & 256$\times$256 & 10 & Yes & Yes & MSI & - & 14 & France \\
MDAS~\cite{mdas} & 6268 & 1,371$\times$888, 1,371$\times$888, 1,170$\times$765 & 0.25, 10 and 30 & No & Yes & HSI, MSI & DSM & 14 & Augsburg, Germany\\
DFC20~\cite{dfc20} & 180,662 & 256$\times$256 & 10 & No & Yes & MSI & - & 10 & Globally distributed  \\
GeoNRW~\cite{geonrw} & 7,782 & 1,000$\times$1,000 & 1 & No & Yes & MSI & DEM & 10 & North-Rhine, Westphalia  \\
PASTIS-R~\cite{partisr} & 2433 times series & 128$\times$128 & 10 & Yes & Yes & MSI & - & 18 & French metropolitan  \\
DFC22~\cite{dfc22} & 5,016 & 2,000$\times$2,000 & 0.5 & No & No & RGB & DEM & L1: 12; L2: 14 & France  \\
FLAIR-one~\cite{flairone} & 77,412 & 512$\times$512 & 0.2 & Yes & No & MSI & DSM, DTM & 13 & French metropolitan  \\
Hunan Multimodal Dataset~\cite{hunanmultimodal} & 450 & 256$\times$256 & 10 & No & Yes & MSI & SRTM & 7 & Hunan, China  \\\hline
CerraData-4MM & 30,322 & 128$\times$128 & 10 & No & Yes & MSI & - & Level-1: 7, level-2: 14 & Bico do Papagaio ecoregion  \\
\hline
\end{tabular}
\end{table*}

\IEEEPARstart{T}{he} Brazilian \textit{Cerrado} is renowned as the most biodiverse savanna in the world\cite{Strassburg}, featuring vegetation highly adapted to prolonged dry seasons. Spanning a significant portion of Brazil{'}s territory, the \textit{Cerrado} is divided into ecoregions, each defined by distinct characteristics shaped by their geographical location and geological formations, contributing to its diverse vegetation \cite{ecorregiao}. In addition, this biome plays an important role in Brazil{'}s agricultural production, accounting for more than 55\% of the nation{'}s food output \cite{santana}. However, land use and land cover (LULC) classification in the \textit{Cerrado} faces unique challenges due to the biome{'}s ecological diversity and the extensive modifications driven by human activities.

The TerraClass program \cite{thematicmap2020,terraclass} plays an essential role in mapping LULC changes in the \textit{Cerrado} region and identifies areas that have undergone deforestation and subsequent conversion to agriculture, pastures, secondary vegetation, and other types of land cover. According to the program, agricultural areas have expanded, primarily at the expense of pasture. However, the establishment of new pasture areas remains the leading driver of natural vegetation conversion in the \textit{Cerrado} biome. These maps support detailed analyses of changes in vegetation cover every two years. However, the manual interpretation-based approach to creating these maps is time-consuming and expensive. To overcome these limitations, machine learning (ML) models offer a promising solution to perform semantic segmentation tasks more efficiently and quickly. However, the \textit{Cerrado} region lacks a substantial labeled dataset suitable for training such models.

Optical images are commonly used to train ML models for such classification tasks, mainly multispectral images \cite{alana}, in order to identify specific visual features of the vegetation. However, optical imagery is limited by factors including illumination, weather, and climatic conditions, such as the presence of clouds or smoke. In contrast, Synthetic Aperture Radar (SAR) technology offers the capability to acquire images of the Earth{’}s surface under all lighting and weather conditions. In addition, this active sensor can penetrate clouds, foliage, and even layers of dry soil. The integration of optical imagery and SAR data has the potential to mitigate the drawbacks of each modality and, most importantly, enhance the representation of LULC patterns \cite{fonseca}.

For other regions, multi-modal datasets of Earth observation imagery paired with semantic reference data are available but focus on distinctively different semantic categories.
MultiSenGE \cite{MultiSenGE} introduces a labeled dataset integrating SAR data from Sentinel-1 and multispectral data from Sentinel-2. It contains 8,157 patches, each measuring 256$\times$256 pixels, collected from the eastern region of France. Additionally, it includes time series data for both modalities. Due to the limited geographical coverage, the dataset{'}s classes predominantly represent urban and agricultural landscapes, such as vineyards and orchards. This specificity renders the dataset highly tailored to a particular application context, with constrained sample diversity and geographic scope.

The SEN12MS \cite{SEN12MS} dataset features globally distributed samples, offering a high diversity of characteristics in 33 classes, with a strong emphasis on natural categories such as various types of forests and savanna formations in all seasons. The dataset has 180,662 patch triplets, including dual-polarization SAR (Sentinel-1), multispectral imagery (Sentinel-2), and MODIS land cover maps. The spatial resolution of the LULC maps (500 meters), as well as the satellite imagery (60, 20, and 10 meters), is resampled to a resolution of 10 meters. The reference maps consequently exhibit low quality in terms of the richness of detail within the polygons.

There are no multimodal benchmark datasets available for the \textit{Cerrado} region. However, optical image sets have been created for scene classification tasks. For example, the Brazilian \textit{Cerrado}-Savanna Scenes dataset \cite{nogueira}. This dataset includes four LULC classes representing the three main groups of natural land cover types and agricultural areas over the preserved region \textit{Serra do Cipó}. The number of available patches is relatively small for effectively training deep learning models, consisting of 1,311 samples. Furthermore, the dataset exhibits a limited representation of class samples due to its restricted diversity. Additionally, the images are limited to capturing only the green, red, and near-infrared bands.

This paper introduces, \textbf{CerraData-4MM} - a multimodal dataset specifically designed for the \textit{Bico do Papagaio} (Parrot{'}s Beak) ecoregion of the \textit{Cerrado}, incorporating SAR and multispectral imagery (MSI) data, along with precise reference masks for LULC classification. The dataset features a two-level hierarchical structure: The first level comprises seven macro classes, while the second level expands into 14 categories, including specific agricultural types, two stages of tree regeneration, and construction types. Additionally, this paper presents the first baseline results using a vision transformer based architecture (ViT) and a U-Net.

\section{Related Datasets}

This section introduces several relevant multimodal datasets published for remote sensing applications. Table~\ref{tab:table1} highlights some of the most state-of-the-art works that incorporate at least two modalities of data, such as SAR and MSI. These datasets differ in scope, spatial resolution, modalities, geographic focus, and classification tasks, addressing various research objectives and applications. Some works offer a rich diversity of samples, as data are collected from different parts of the world, such as SEN12MS and DFC2022, making them suitable for general-purpose studies. In contrast, others are designed for regional and specific areas of interest, providing high-resolution data for localized analyses, such as GeoNRW and the Hunan Multimodal Dataset.

In particular, MultiSenGE, PASTIS-R, and FLAIR-one incorporate time series data, enabling temporal analysis. In contrast, SEN12MS and DFC20 consist solely of single snapshot imagery. Additionally, MDAS and DFC22 utilize complementary data, such as digital surface models (DSM) and digital terrain models (DTM), alongside SAR and MSI images. Despite the availability of various data modalities that support model development for diverse remote sensing applications, many datasets primarily focus on land use classes, such as buildings, airports, and roads. SEN12MS is a notable exception, offering a comprehensive list of 33 classes related to vegetation types, including urban areas and vegetation domains.


In general, the primary factor that significantly restrains these datasets is concerning the quality and accuracy of the reference data, e.g. SEN12MS. In this context, another relevant aspect is related to the focus on land use classes, particularly regarding the dataset{'}s heterogeneity for studies aiming to assess the transferability potential of such reference data to other study areas. CerraData-4MM provides accurate reference data designed for a spatial resolution of 10 meters, as well as a wide heterogeneity of land use and land cover class samples, despite being focused on only one ecoregion of \textit{Cerrado}.

The \textit{Cerrado} Dataset (CerraData) \cite{cerradata} is a comprehensive resource designed to advance research in land use and land cover (LULC) scene classification by providing labeled satellite imagery for deep learning (DL) approaches \cite{fewshot}. CerraData-1 includes 1.5 million unlabeled images collected from approximately 44\% of the \textit{Cerrado} biome, serving as a robust foundation for exploratory analyses. CerraData-2 introduces 50,000 labeled samples categorized into five LULC classes, while CerraData-3 expands the dataset further with 80,000 labeled samples, enhancing its diversity and applicability. All datasets are obtained from the CBERS-4A satellite, which provides imagery in near-infrared (NIR), green, and blue bands with a spatial resolution of 2 meters. The labeled datasets, CerraData-2 and CerraData-3, are specifically designed for scene classification tasks.

CerraData-4MM, in contrast to other datasets that are either globally distributed or narrowly focused on a small area of interest such as a single city, encompasses the \textit{Bico do Papagaio} ecoregion within the \textit{Cerrado} biome. This design promotes the development of regional studies and applications aimed at conserving and managing this unique and biodiverse biome, which is found in a transitional area with the Amazon forest. The dataset comprises 30,291 image patches, each 128$\times$128 pixels with a spatial resolution of 10 meters, and combines a substantial volume of accurately labeled data, enriched by its hierarchical organization into two levels of classes. CerraData-4MM is particularly challenging at the second class level, especially regarding vegetation regeneration stages and types of agriculture since the categories present similar patterns.

\section{The dataset}

\begin{figure}[!t]
\centering
\includegraphics[width=3.3in]{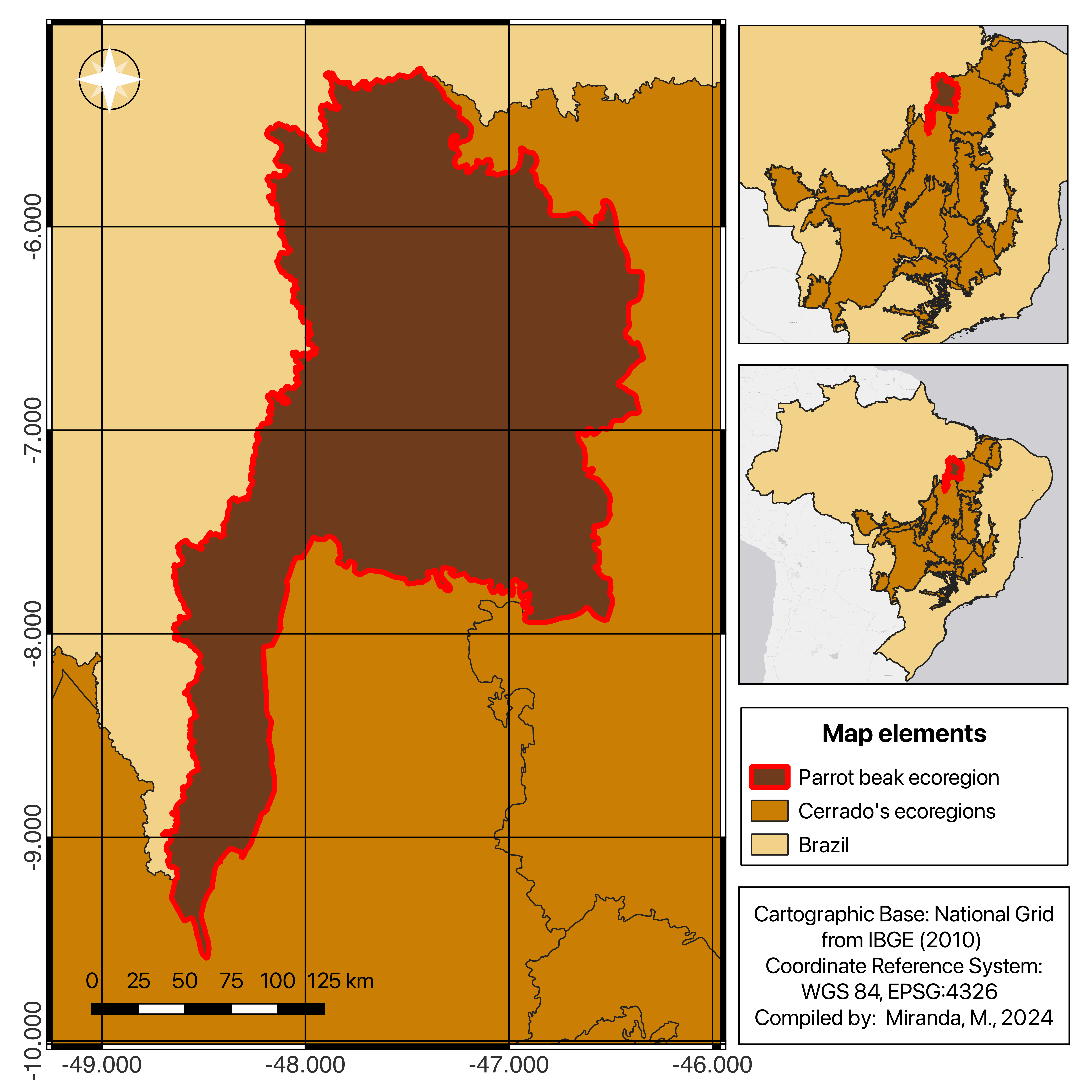}
\caption{The \textit{Bico do Papagaio}, marked in brown with red edges, is the region selected to create the dataset.}
\label{fig:map}
\end{figure}

CerraData-4MM is a multimodal benchmark dataset crafted for the \textit{Cerrado} biome. Among the nineteen ecoregions encompassing the \textit{Cerrado} \cite{ecorregiao}, it specifically targets the \textit{Bico do Papagaio}. As depicted in Figure~\ref{fig:map}, this region is situated in the northernmost expanse of Tocantins state and adjoins the Amazon rainforest. The unique confluence of high biodiversity within its natural vegetation and extensive agricultural activities has resulted in approximately 25\% of the area being occupied by anthropogenic activities \cite{santana}. It stands out as one of the least populated regions within the \textit{Cerrado} and it retains a remarkable proportion of natural land cover, underscoring its ecological richness and conservation significance.

\subsection{The data source}

CerraData-4MM integrates SAR and multispectral imagery from the Sentinel-1 (S1) and Sentinel-2 (S2) satellites, respectively. The images are collected in 2022 to temporally align with the reference data (see Section~\ref{sec:refdata}). Days with minimal cloud cover are prioritized to ensure optimal visibility of the targets. For S1, seven scenes are used, each featuring dual-polarization (VV and VH) bands from the Level-1 Ground Range Detected (GRD) product. These bands have a spatial resolution of 10 meters. For S2, 18 scenes are used. Each scene includes four bands with a 10 meters spatial resolution (B2, B3, B4, and B8), six bands with a 20 meters spatial resolution (B5, B6, B7, B8a, B11, and B12), and two atmospheric bands with a 60 meters spatial resolution (B9 and B10).

\subsection{Reference data}
\label{sec:refdata}

The reference data are based on a land use and land cover map produced by the TerraClass \textit{Cerrado} program in 2022. This map categorizes the land into 15 classes, including primary and secondary natural vegetation, five types of agricultural land, water bodies, and deforested areas. This highly accurate map is meticulously crafted by experts in LULC classification. It is underpinned by Sentinel-2 imagery, offering a spatial resolution of 10 meters. The classification process entails manual categorization of classes pertaining to urbanization (urban areas, other built structures, mining, and other uses) and time series analysis employing deep learning models for classes related to agriculture (temporary agriculture for a single cycle, temporary agriculture for multiple cycles, perennial and semi-perennial agriculture). Furthermore, Self-Organized Maps (SOMs) are utilized to identify secondary natural vegetation and pasture classes.

\subsection{Data preprocessing}

\begin{table}[!ht]
\caption{Sampling distribution by class.\label{tab:table2}}
\raggedright
\begin{tabular}{lc >{\raggedright\arraybackslash}p{2cm} lc}
\hline
\textbf{L1} & \textbf{\%} & \textbf{L2} & \textbf{\%} & Samples\\
\hline
Pasture (Pa)    &   35.25   &   Pasture (Pa)    &   35.25   & 174929011 \\
\hline
Arboreal (Ab)   &   58.43   &   Primary Natural Vegetation (V1) & 52.21  & 259126072 \\ 
                &           &   Secondary Natural Vegetation (V2) & 6.22  & 30874079 \\
\hline
Agriculture (Ag)&   4.43    &   Perennial (Pr)  &    0.07   &   371519 \\
                &           &   Semi-perennial (SP) & 0.15  &   767292\\
                &           &   Temporary agriculture of 1 cycle (T1) & 0.27  & 1348807\\
                &           &   Temporary agriculture of 1+ cycle (T1+) & 1.15  & 5702780\\
                &           &   Forestry (Ft)   &   1.49    &   7388637 \\
                &           &   Deforestation (Df) &  1.30  &  6466058\\
\hline
Mining (Mg)     &   0.01    &   Mining (Mg) &   0.01    &   63964\\
\hline
Building (Bg)   &   0.55    &   Urban area (UA) & 0.45  &   2241073 \\
                &           &   Other Built area (OB) & 0.10  &  508261\\
Water body (Wt)      &   1.32    &   Water body (Wt) & 1.32  &  6447232\\
\hline
Other Uses (OU) &   0.01    &   Other Uses (OU) & 0.01  & 52959 \\
\hline
\end{tabular}
\end{table}

Each SAR and MSI scene is processed using the SNAP tool\footnote{SNAP tool: \url{https://step.esa.int/main/toolboxes/snap/}}. All bands are resampled to spatial resolution of 10 meters. The bands are then stacked to generate rasters comprising 12 channels for optical images and 2 channels for SAR images. The scenes are re-projected to the EPSG 4326 coordinate system using bilinear interpolation and cropped into patches with dimensions of 128$\times$128 pixels. Patches containing NaN or Null values are discarded.

The reference data is organized into two hierarchical levels of classes, as illustrated in Figure~\ref{fig:dataset}. 
At the first level, seven macro categories (L1) are considered, while the second level (L2) comprises 14 classes based on the TerraClass 2022 map. As outlined in Table~\ref{tab:table2}, the primary distinction between these levels lies in the grouping of subclasses into broader categories. For instance, the ``Agriculture” class in L1 is subdivided into five types of agricultural management in L2: ``Perennial Agriculture”, ``Semi-perennial Agriculture”, ``Temporary agriculture of 1 cycle”, ``Temporary agriculture 1+ cycle”, and ``Forestry”.

\subsection{CerraData-4MM}

\begin{figure}[!ht]
\centering
\includegraphics[width=3.5in]{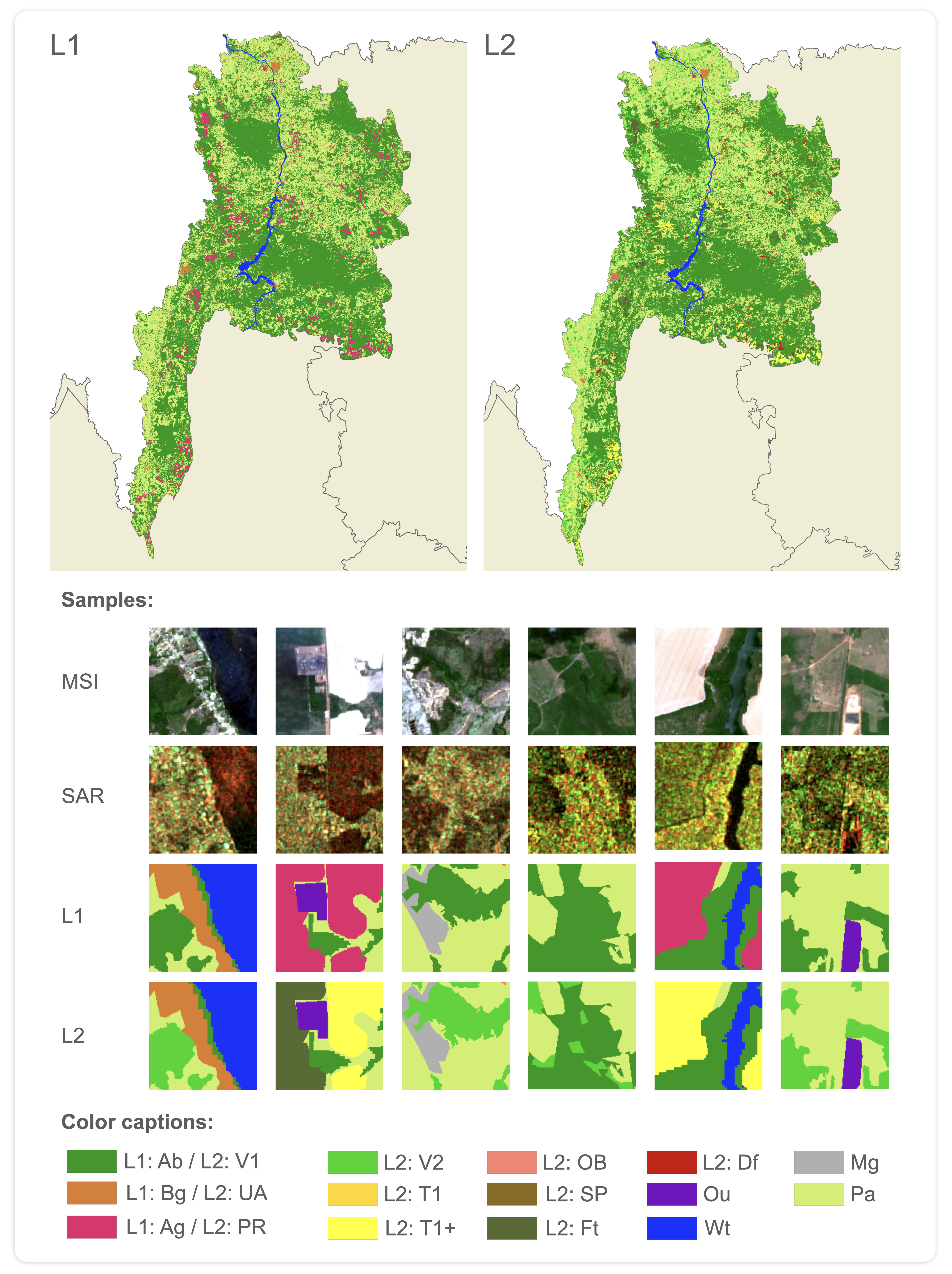}
\caption{CerraData-4MM{'}s hierarchical level of classes and its samples.}
\label{fig:dataset}
\end{figure}

CerraData-4MM comprises sets of SAR and MSI data, followed by semantic maps for L1 and L2, respectively. The two class levels create a diverse and challenging dataset that covers categories of regeneration level, deforestation increment, as well as different types of agriculture. Each set contains 30,322 patches with a spatial resolution of 10 meters. This dataset provides a rich diversity of classes representative of the \textit{Bico do Papagaio} ecoregion. As illustrated in Figure~\ref{fig:dataset}, L2 introduces five additional subcategories of agricultural types, one additional category for built areas, and two distinct vegetation generation classes. Table~\ref{tab:table2} details the distribution of these classes, highlighting the significant challenge of classifying categories with a low percentage of samples at both levels. 
In addition, certain classes exhibit similar visual patterns, such as agriculture and deforestation categories, further complicating classification. Another challenging example is the ``Other Uses" category, which not only has one of the lowest distribution percentages but also presents heterogeneous patterns due to its inclusion of diverse targets.

\section{Evaluation}
To assess the learnability of the dataset, this study employs the baseline architectures U-Net \cite{ronneberger} and TransNuSeg \cite{hetransnuseg}. These models adopt distinct approaches: U-Net operates as a single-task learning (STL) model based on convolutional neural networks (CNN), whereas TransNuSeg functions as a multi-task learning (MTL) model utilizing Vision Transformers (ViT). While U-Net is a well-established model extensively applied in remote sensing tasks, TransNuSeg represents a recently introduced framework specifically designed for semantic segmentation, edge segmentation, and edge nucleus segmentation tasks, primarily within the field of medical imaging.

Considering the CerraData-4MM reference data, adjustments are made to the TransNuSeg architecture, defining the tasks of semantic segmentation and edge detection. Regarding U-Net, its original structure remains unchanged. Given the dataset modalities, the models are trained in three scenarios for both L1 and L2: (i) using only MSI data, (ii) using only SAR data, and (iii) using both modalities in a concatenated approach. No preprocessing is applied to the images; however, class weights derived from the sample distribution (see Table \ref{tab:table2}), calculated based on the reference masks, are used to ensure dataset balance. Model performance is evaluated using the macro-averaged F1-score and mean Intersection over Union (mIoU) metrics. Table \ref{tab:table3} provides additional details regarding the model configurations.

\begin{table}[!ht]
\caption{Models settings\label{tab:table3}}
\centering
\begin{tabular}{>{\raggedright\arraybackslash}p{2cm} >{\raggedright\arraybackslash}p{2cm} >{\raggedleft\arraybackslash}p{3cm}}
\hline
\textbf{Model} & \textbf{Hyperparameter} & \textbf{Setting}\\
\hline
TransNuSeg & Loss function & Dice and Cross Entropy\\
 & Sharing ratio & 0.8\\
U-Net & Loss function & Cross Entropy\\
 & Encoder & ResNet-50\\
All & Optimizer & Adam \\
 & Learning rate & 0.001\\
 & Batch size & 32\\
 & Epochs & 100\\
\hline
\end{tabular}
\end{table}

\subsection{Baseline results}
\begin{figure*}[!tb]
\centering
\includegraphics[width=\textwidth]{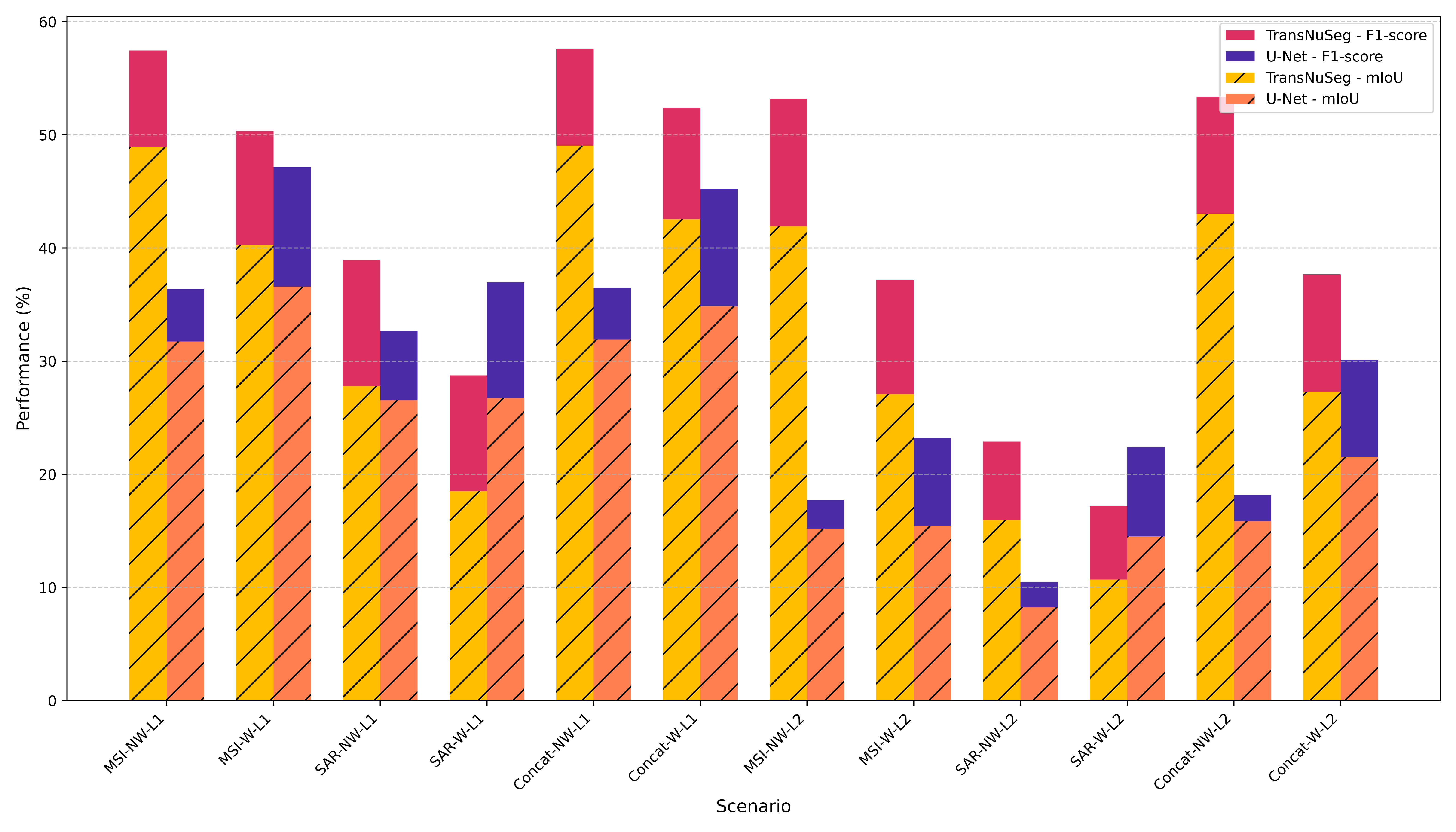}
\caption{Overall Performance (F1-score and mIoU).}
\label{fig:scores}
\end{figure*}

Figure \ref{fig:scores} presents an overview of the overall results for each baseline model across the diverse scenarios. It incorporates scores that indicate the utilization (W) and non-utilization (NW) of class weights for dataset balancing. For instance, the notation ``MSI-NW-L1” denotes the modality set without class weights at the initial hierarchical level of classes. Figure~\ref{fig:scores} shows that TransNuSeg consistently outperforms U-Net in almost all scenarios, mostly at the L1 hierarchical level, whereas U-Net presents significantly lower scores. The ViT-based model exhibits better ability to learn from multimodal data and challenging segmentation tasks of L1 classes.  Both models show medium scores for both metrics.

In the first scenario, where models are trained exclusively on MSI data, the performance is consistently and significantly superior compared to training only on SAR data, for both TransNuSeg and U-Net. The limited performance of models trained on SAR images can be attributed to the inherently noisier nature of SAR data, but also to its properties, which do not emphasize the features of some classes. In contrast, the MSI dataset comprises 12 bands, providing rich visual features of the targets. The concatenation of MSI and SAR modalities markedly enhances the results compared to the use of either modality individually. 
For instance, in the Concat-NW-L1 scenario, the ViT-based model achieves an F1-score of 57.60\%, whereas for the ``MSI-NW-L1” and ``SAR-NW-L1” scenarios, the corresponding scores are 57.45\% and 38.94\%, respectively.

The models{'} scores drop considerably for the second-level classes. The increased complexity of L2 is attributed to factors such as the greater number of classes, the uneven distribution of classes, spatial resolution, and inter-class correlations. For instance, in the ``MSI-NW" scenario, TransNuSeg achieves an F1-score of 57.45\% and an mIoU of 48.94\% in L1, but these metrics decline to 53.17\% and 41.90\%, respectively, in L2. U-Net exhibits even more unsatisfactory performance; in the ``Concat-NW-L2" scenario, its F1-score and mIoU decrease to 18.16\% and 15.84\%, respectively.

With regards to weights-based dataset balancing, this technique results in an overall reduction in F1 and mIoU scores across several scenarios for the ViT-based network, as observed in ``MSI-W-L1" compared to ``MSI-NW-L1". This outcome indicates that the weighted adjustment employed to address class imbalance may adversely affect the performance of the more represented classes, highlighting the sensitivity of TransNuSeg to this technique. U-Net exhibits improvements in specific cases, such as ``SAR-W-L1" compared to ``SAR-NW-L1".

\begin{figure}[!htb]
\centering
\includegraphics[width=3.3in]{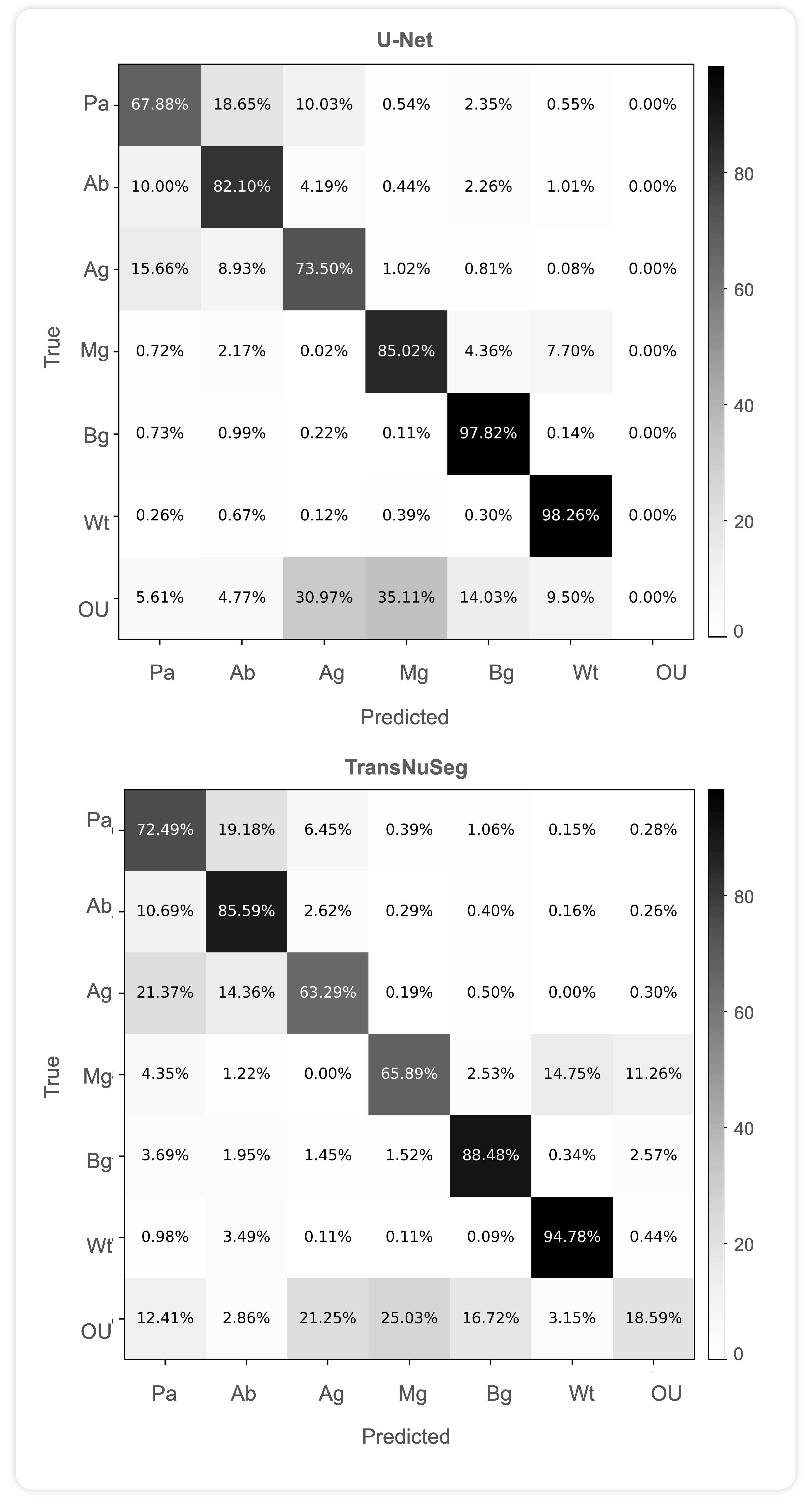}
\caption{Confusion matrix for weighted MSI+SAR L1 data-trained models.}
\label{fig:l7}
\end{figure}

\begin{figure}[!ht]
\centering
\includegraphics[width=3.3in]{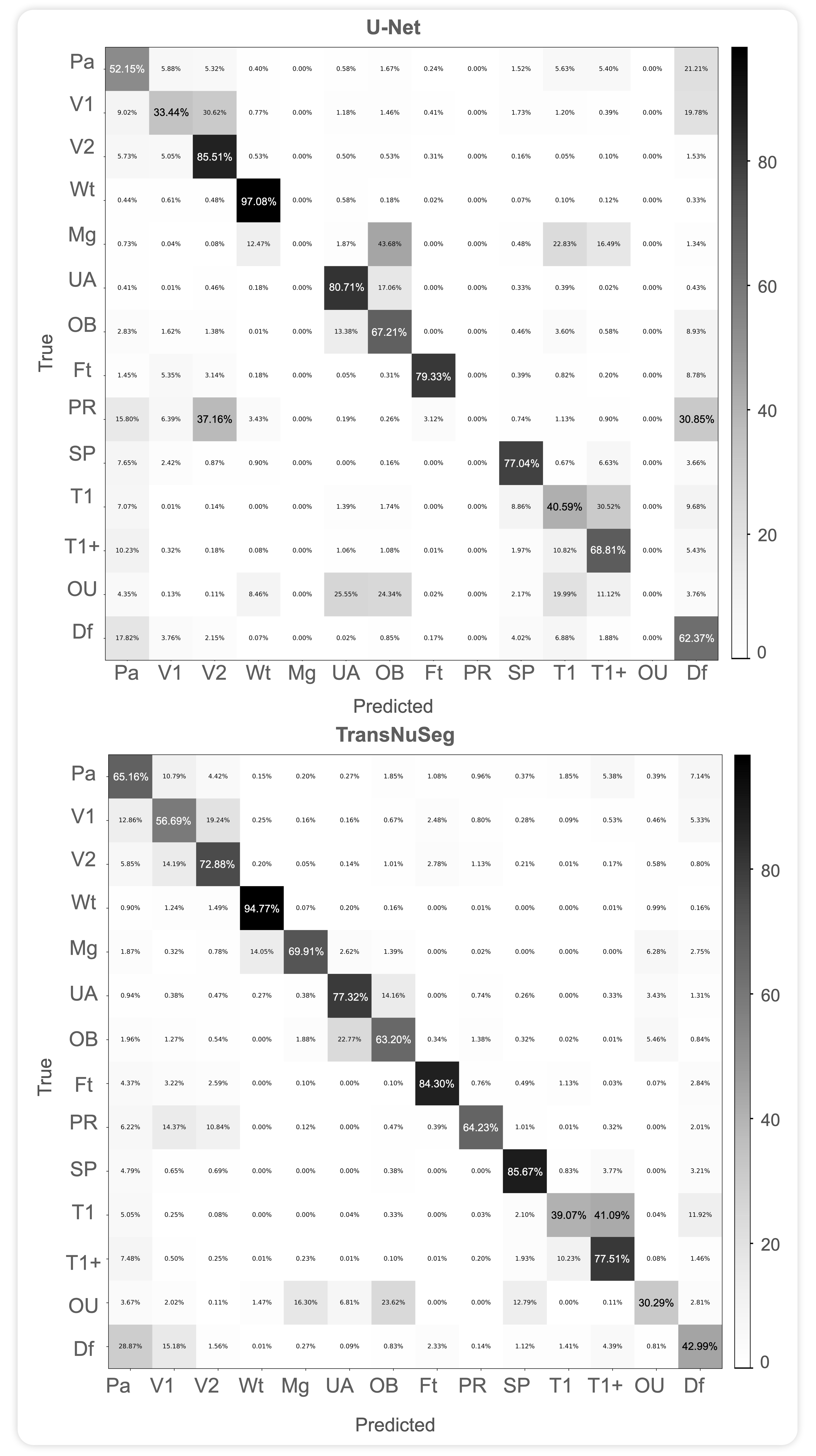}
\caption{Confusion matrix for weighted MSI+SAR L2 data-trained models.}
\label{fig:l14}
\end{figure}

Given the performance improvement observed when models are trained on both modalities, Figures \ref{fig:l7} and \ref{fig:l14} show the confusion matrix of both U-Net and TransNuSeg at the two different hierarchical levels. Although the overall performance of models trained without dataset balancing is slightly better than when trained with balancing, the confusion matrices demonstrate a tangible improvement with the use of weights assigned to classes with a smaller sample distribution. The reduction in the F1-score and mIoU occurs because balancing with weights diminishes the bias in favor of the majority classes, leading to broadly balanced performance among all classes. Nonetheless, the cost of this approach is a slight loss of accuracy in the dominant classes, which affects the metric scores.

TransNuSeg slightly outperforms U-Net, demonstrating its greater generalization ability. For intermediate classes, U-Net outperforms TransNuSeg in accuracy, especially on the ``Water" class, which is highly distinct and easy to segment. ``Agriculture", on the other hand, suffers a more from confusion in TransNuSeg, indicating that the model may have difficulty capturing subtle details between ``Agriculture" and dominant classes, such as ``Pasture". Minority classes continue to be the biggest challenge for both models. U-Net excels in some of these classes, such as ``Building", but both models struggle to identify ``Mining" and ``Other Uses", which is expected given the lack of representation in training.

Figure \ref{fig:l14} presents the confusion matrices for the models trained on the second hierarchical level of CerraData-4MM with class balancing. As highlighted in Table \ref{tab:table2}, the sample distribution is highly imbalanced, with classes such as V1 and Pa being predominant, while others like Mg, OB, T1, T1+, and OU are underrepresented. This imbalance leads to a high rate of misclassification among similar or minority classes, negatively impacting overall F1 and mIoU scores, as detailed in Table \ref{tab:table3}. Despite these challenges, balancing the dataset facilitates the models in learning about the minority classes. However, even with the application of class weights during training, the models continue to struggle to generalize effectively to underrepresented and similar classes, such as V1, V2, and Df.

\begin{figure}[!ht]
\centering
\includegraphics[width=3.3in]{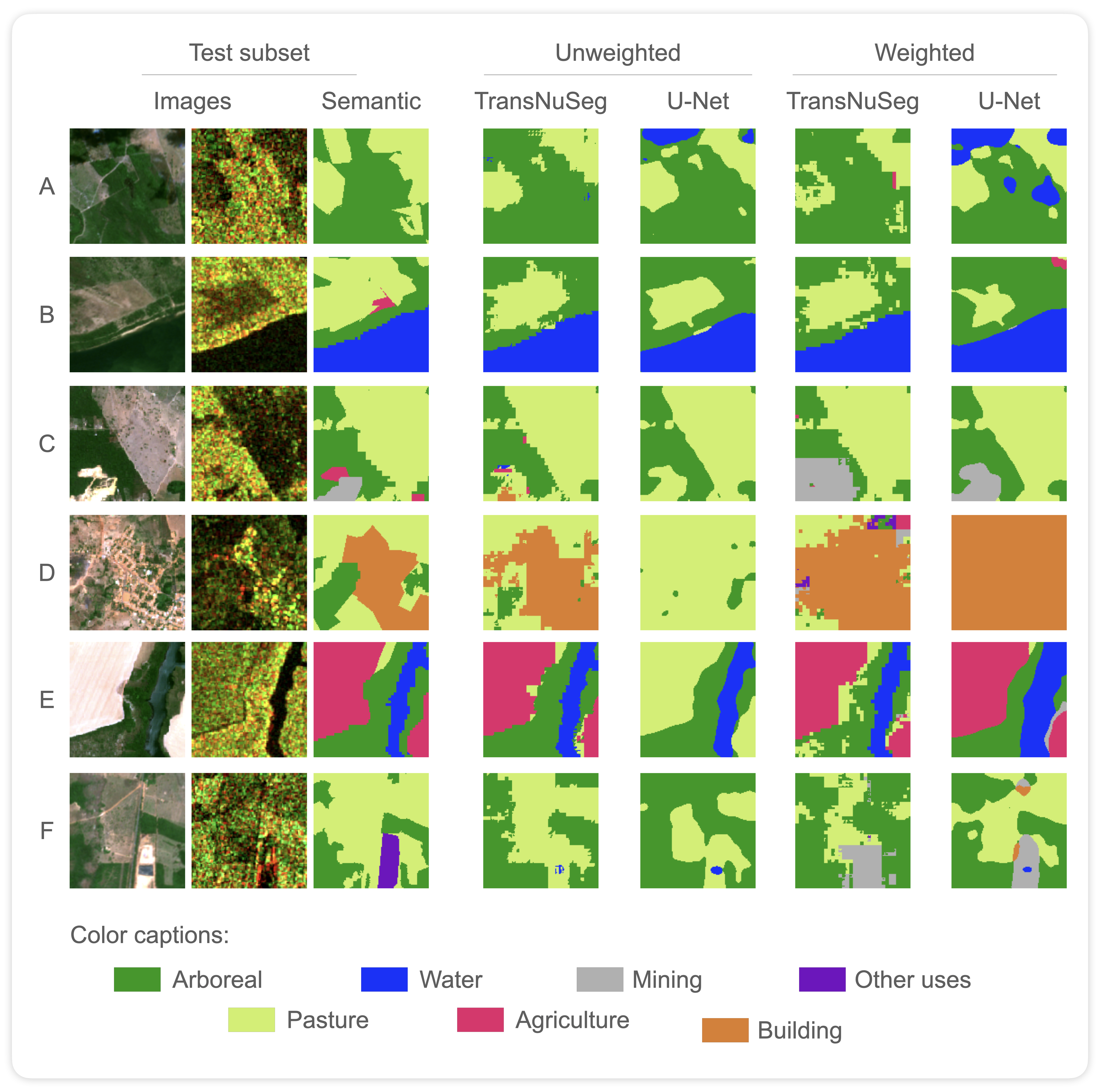}
\caption{Samples of predictions by U-Net and TransNuSeg trained with both modalities in L1.}
\label{fig:pred_l7}
\end{figure}

In Figure \ref{fig:pred_l7}, most classes, including Ab and Pa, consistently predict the same values, especially for scenarios where weights are assigned to classes. Using weighted data leads to substantial improvements in detecting underrepresented classes, such as Bg, Wt, and Mg. Therefore, this technique mitigates the bias towards dominant classes like Pa in the TransNuSeg model, as seen in cases D and E. The TransNuSeg typically produces more refined segmentations with reduced boundary dispersion, demonstrating its ability to capture relevant details in homogeneous regions. On the other hand, the U-Net model exhibits higher confusion between Pa and Ag, particularly in the unweighted scenario. This issue is particularly noticeable in row E, where agricultural areas are frequently misclassified as pasture.

\begin{figure}[!ht]
\centering
\includegraphics[width=3.3in]{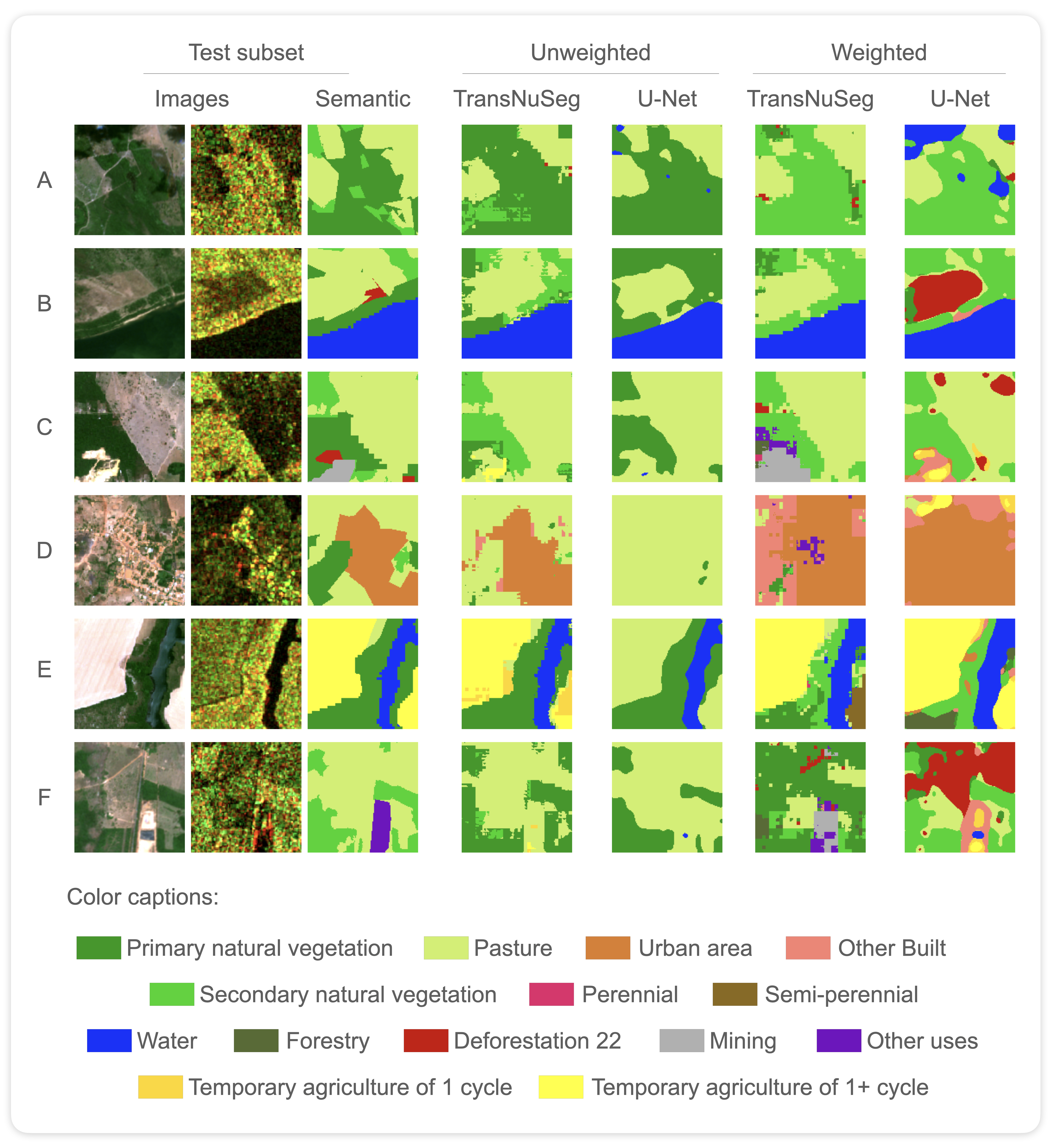}
\caption{Samples of predictions by U-Net and TransNuSeg trained with both modalities in L2}
\label{fig:pred_l14}
\end{figure}

In the models trained in L2 (see Figure \ref{fig:pred_l14}), the first notable aspect is the increase in classification errors observed between samples A and B when comparing the unweighted and weighted training approaches. These examples underscore the trade-off introduced by applying class weights to emphasize minority classes. Despite this trade-off, the representation of minority classes improves significantly, as evidenced in example F, where the OU class is successfully identified by TransNuSeg. Moreover, the ViT-based model demonstrates superior capability in capturing classes such as Df, OU, and PR. In contrast, while U-Net is also able to identify these classes, its predictions tend to be more fragmented and exhibit higher levels of visual noise.

In the case without class balancing, both models demonstrate a consistent ability to identify majority classes while struggling to detect minority classes. TransNuSeg produces smoother and more generalized results compared to U-Net, which tends to exhibit spatial fragmentation in some areas of its predictions. Given that TransNuSeg{'}s architecture is based on vision transformers, i.e., attention mechanisms, it is inherently capable of distributing attention more evenly across all classes, unlike U-Net. Consequently, U-Net often overlooks local variations and fills large areas with the dominant class. This is particularly evident in samples D and E, where T1+ or UA are misclassified as Pa or V1.

\section{Final remarks}
This paper introduces CerraData-4MM, a multimodal dataset developed for land use and land cover classification in the \textit{Cerrado} biome. The dataset integrates SAR and MSI imagery from Sentinel-1 and Sentinel-2, with patches at a spatial resolution of 10 meters. It provides two hierarchical levels of classification, comprising 7 and 14 categories, along with corresponding edge information. CerraData-4MM encompasses the \textit{Bico do Papagaio} ecoregion, which poses unique classification challenges due to its mosaic of natural vegetation, anthropogenic activities, and agricultural subcategories.
This dataset presents a real-world challenge due to the prevalence of class imbalance, which may require pre-processing techniques and models capable of learning from imbalanced, heterogeneous, and visually ambiguous data.

The experimental results underscore these challenges while also highlighting the opportunities for innovation in deep learning (DL) models. Approaches such as U-Net and TransNuSeg demonstrate strong performance in majority classes but struggle with minority categories and subtle spectral differences within agricultural subclasses. The weighted dataset strategy improves the representation of underrepresented classes, showcasing the dataset's utility in testing class-balancing techniques and model adaptations.
However, consistent misclassifications in visually similar classes emphasize the necessity for advanced techniques, such as attention mechanisms, multimodal fusion, and hierarchical learning frameworks.

For the next version of CerraData, it is expected to include other Cerrado ecoregions in order to further increase the diversity of samples, with regard to visual patterns. Given the diversity of vegetation domains, it is planned to include more natural classes, e.g., rupestrian savannas and gallery forests, thus encouraging the mapping and historical reporting of environmental preservation of such groups of vegetation formation. The integration of other modalities, e.g., DEM, for height estimation tasks can be considered to promote support for multi-task learning models.

\section*{Acknowledgments}

This research was conducted as part of the project \textit{Classificação de imagens e dados via redes neurais profundas para múltiplos domínios} (Image and data classification via Deep neural networks for multiple domainS -- \textbf{IDeepS}). The project repository is publicly available online at \href{https://github.com/vsantjr/IDeepS}{https://github.com/vsantjr/IDeepS}.
The authors thank the São Paulo Research Foundation (FAPESP, grant no 2023/09118-6), the Brazilian National Council for Scientific and Technological Development (CNPq, grant no \#302205/2023-3, \#140459/2023-5), Institutional Program for Internationalization by the Brazilian Federal Agency for Support and Evaluation of Graduate Education (PrInt-CAPES, grant \#88887.936146/2024-00) for their support of a research exchange at the German Aerospace Center (DLR) in Oberpfaffenhofen, for which we are especially grateful. 
We also appreciate the support of the \emph{Laborat{\'o}rio Nacional de Computa\c{c}\~{a}o Cient{\'i}fica} (LNCC---National Laboratory for Scientific Computing, MCTI, Brazil) for providing access to the computational resources of the SDumont supercomputer.

\bibliographystyle{IEEEtran}
\bibliography{cerradata4mm_paper}


\newpage

\section{Biography Section}

\begin{IEEEbiography}[{\includegraphics[width=1in,height=1.25in,clip,keepaspectratio]{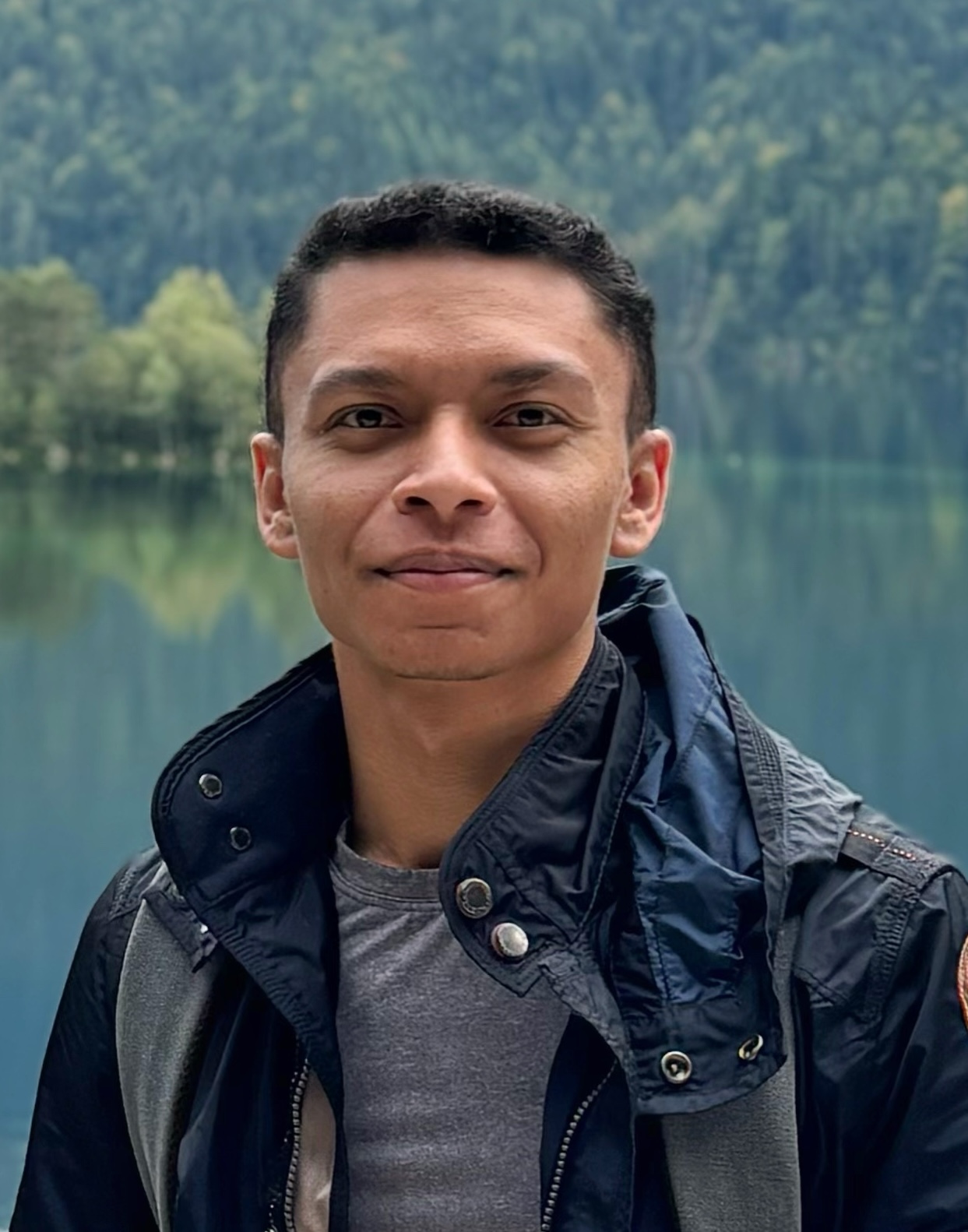}}]{Mateus de Souza Miranda}
is a PhD student in Applied Computing at the National Institute for Space Research (INPE), where he also earned his Master's degree in the same field. His research specializes in multi-task learning and self-supervised deep learning models applied for land use and land cover classification through satellite images.

\end{IEEEbiography}

\begin{IEEEbiography}
[{\includegraphics[width=1in,height=1.25in,clip,keepaspectratio]{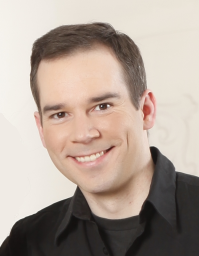}}]{Ronny Hänsch}
is a post-doctoral researcher and serves as the lead of the "Machine Learning" team in the Department of SAR Technology at the German Aerospace Center (DLR). His research focuses on advanced machine learning techniques, including deep learning and random forests, applied to remote sensing with a particular emphasis on Polarimetric Synthetic Aperture Radar (PolSAR) imagery. Additionally, his work explores innovative domains such as self-supervised learning and cross-modal learning.

\end{IEEEbiography}

\begin{IEEEbiography}
[{\includegraphics[width=1in,height=1.25in,clip,keepaspectratio]{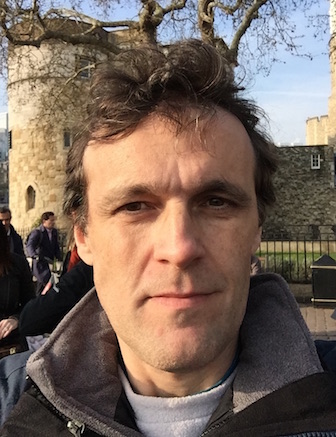}}]{Valdivino Alexandre de Santiago Júnior} holds a Ph.D. from the Postgraduate Program in Applied Computing (2011) of the \textit{Instituto Nacional de Pesquisas Espaciais} (INPE) in Brazil and a Master's Degree in Electrical Engineering from the \textit{Universidade Federal do Ceará} (1999) in Brazil. He was a visiting researcher in Computer Science at the University of Nottingham, England, United Kingdom (2019) and in Formal Verification of Probabilistic Systems at Concordia University, Montreal, Canada (2015). He also holds the position of senior technologist and is the leading researcher at the \textit{Laboratório de Inteligência ARtificial para Aplicações AeroEspaciais e Ambientais} (LIAREA) at INPE. He worked for over 20 years in the development of scientific satellites and stratospheric balloon projects. Research topics of interest include artificial intelligence, machine learning, deep learning, large language models, remote sensing, environmental applications, systems/software optimization, and aerospace systems (satellites, drones).

\end{IEEEbiography}

\begin{IEEEbiography}
[{\includegraphics[width=1in,height=1.25in,clip,keepaspectratio]{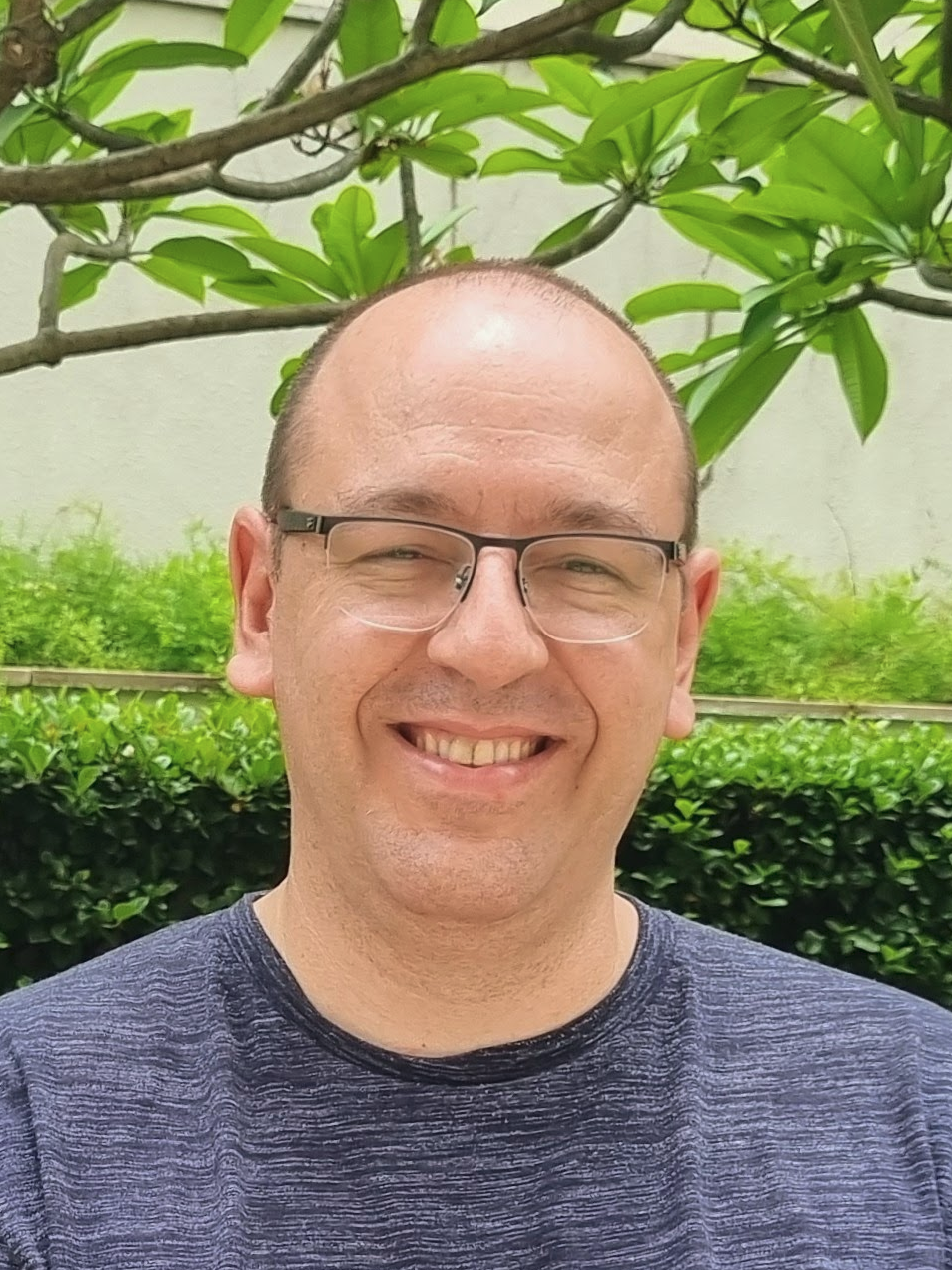}}]{Thales Sehn Körting}
holds a Ph.D. in Remote Sensing and an M.S. in Applied Computing, both from Brazil's National Institute for Space Research (INPE), as well as a B.S. in Computer Engineering from FURG. His research interests include remote sensing image segmentation, multi-temporal analysis, image classification, data mining algorithms, and artificial intelligence.

\end{IEEEbiography}

\begin{IEEEbiography}
[{\includegraphics[width=1in,height=1.25in,clip,keepaspectratio]{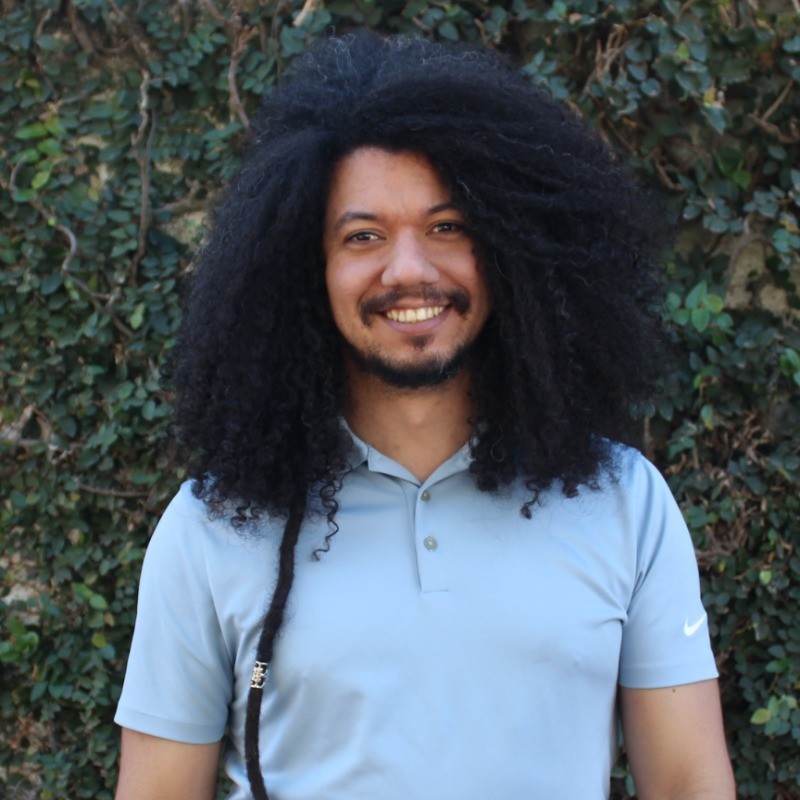}}]{Erison Carlos dos Santos Monteiro}
I grew up on a small farm in the interior of São Paulo State, near the Paraná River, where I learned about the complexity of life. During my undergraduate studies in biology, I discovered how science can be used to explain the world.
In my PhD in Ecology and Biodiversity at São Paulo State University (UNESP) and as a visiting researcher at the Senckenberg Biodiversity and Climate Research Centre (SBK-F) in Frankfurt, Germany, I investigated how human activities on landscapes affect species interactions and how the extinction of these interactions can influence the evolutionary aspects of biodiversity.
Recently, I was appointed to a technician position at the Brazilian National Institute for Space Research (INPE), where I coordinate the INPE team for TerraClass \textit{Cerrado}—a project mapping the land cover of the \textit{Cerrado} biome in Brazil.
Currently, I am interested in playing rugby in the 3rd division of the São Paulo State Championship and exploring how human activities drive land cover change dynamics, alter landscapes, and impact biodiversity and ecosystem services through the use of remote sensing and computational approaches.

\end{IEEEbiography}

\vfill

\end{document}